\documentclass{article}
\usepackage{ijcai22}

\usepackage{times}
\usepackage{soul}
\usepackage{url}
\usepackage[hidelinks]{hyperref}
\usepackage[utf8]{inputenc}
\usepackage[small]{caption}
\usepackage{graphicx}
\usepackage{amsmath}
\usepackage{amsthm}
\usepackage{booktabs}
\usepackage{algorithm}
\usepackage{algorithmic}
\usepackage{xspace}
\usepackage{subfigure}
\usepackage{dcolumn}
\newcolumntype{d}[1]{D{.}{.}{#1}}

\newcommand{\eat}[1]{}

\renewcommand{\vec}[1]{\mathbf{#1}}

\newcommand{\eg}{\emph{e.g.,}\xspace}

\title{Automatic Noisy Label Correction for Fine-Grained Entity Typing}

\author{
Weiran Pan$^{1,2}$
\and
Wei Wei$^{1,2}$\footnote{
    Corresponding author.
}
\and
Feida Zhu$^3$
\affiliations
$^1$Cognitive Computing and Intelligent Information Processing (CCIIP) Laboratory, School of Computer Science and Technology, Huazhong University of Science and Technology, China\\
$^2$Joint Laboratory of HUST and Pingan Property \& Casualty Research (HPL), China\\
$^3$School of Computing and Information Systems, Singapore Management University, Singapore\\
\emails
\{panwr, weiw\}@hust.edu.cn,
fdzhu@smu.edu.sg
}

\begin{document}

\maketitle

\begin{abstract}

Fine-grained entity typing (FET) aims to assign proper semantic types to entity mentions according to their context, which is a fundamental task in various entity-leveraging applications. Current FET systems usually establish on large-scale weakly-supervised/distantly annotation data, which may contain abundant noise and thus severely hinder the performance of the FET task. Although previous studies have made great success in automatically identifying the noisy labels in FET, they usually rely on some auxiliary resources which may be unavailable in real-world applications (\eg pre-defined hierarchical type structures, human-annotated subsets). In this paper, we propose a novel approach to automatically correct noisy labels for FET without external resources. Specifically, it first identifies the potentially noisy labels by estimating the posterior probability of a label being positive or negative according to the logits output by the model, and then relabel candidate noisy labels by training a robust model over the remaining clean labels. Experiments on two popular benchmarks prove the effectiveness of our method. Our source code can be obtained from \url{https://github.com/CCIIPLab/DenoiseFET}.

\end{abstract}

\section{Introduction}
Fine-grained entity typing (FET) \cite{ling2012fine} aims to predict fine-grained semantic types for entity mentions according to their contexts. Recently, \cite{choi2018ultra} further proposed the ultra-fine entity typing introducing a richer type set. The fine-grained entity types obtained by FET is beneficial for many downstream NLP tasks like
entity linking \cite{onoe2020fine}, relation extraction \cite{koch2014type,shang2020noisy,shang2020learning}, question answering \cite{wei2011integrating,wei2016exploring}, dialogue \cite{wei2019emotion,wei2021target} and coreference resolution \cite{onoe2020interpretable}.

\begin{figure}[t]
    \centering
    \includegraphics[width=\linewidth]{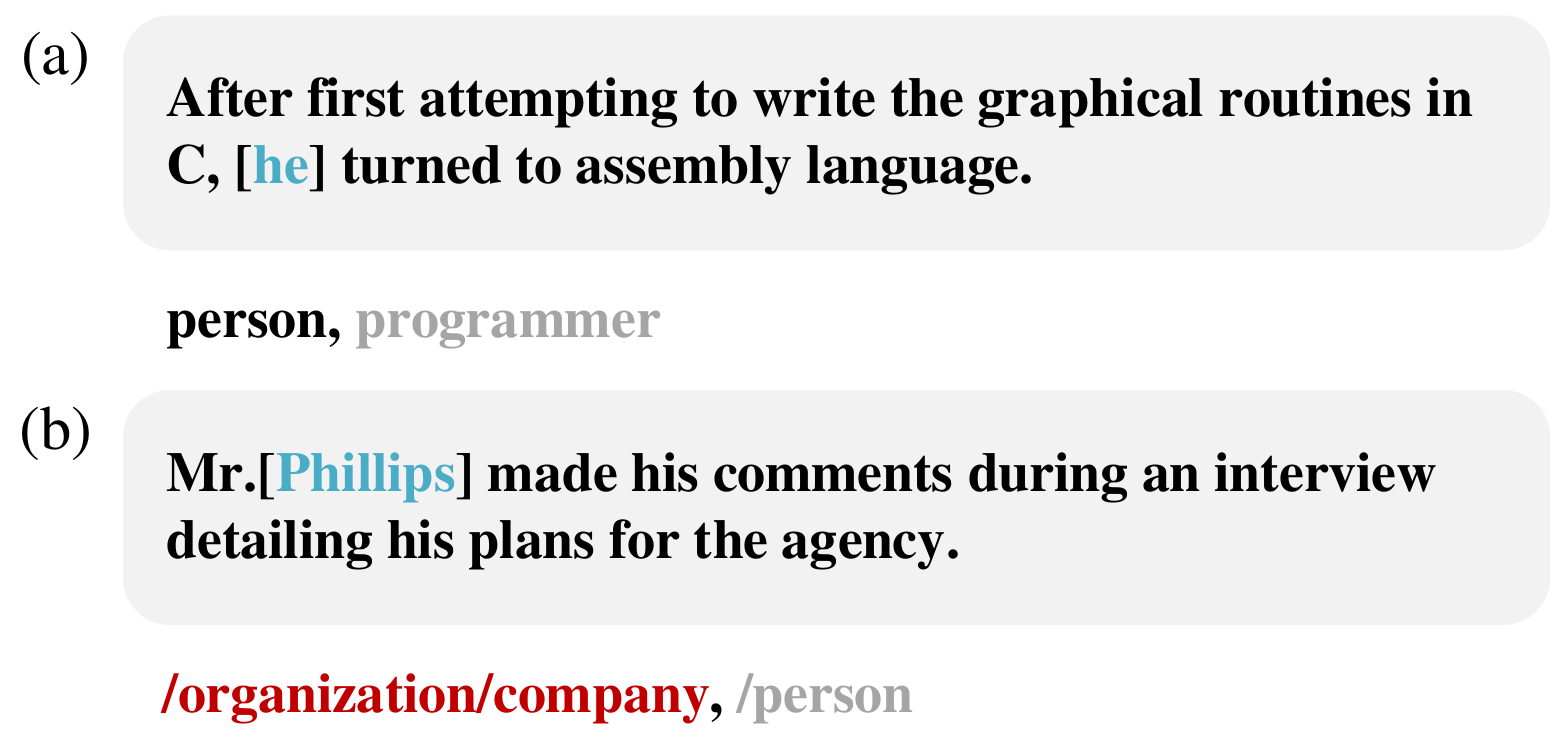}
    \caption{Noisy label examples, the false-positive label is present in red, false-negative labels are present in grey. (a): A manually-annotated example selected from the Ultra-Fine dataset. The annotator failed to cover all types. (b): A distantly-labeled example chosen from the OntoNotes dataset. Context-agnostic distant supervision generated mislabeled sample.}
    \label{fig:noisylabels}
\end{figure}

A fundamental challenge of FET is handling the noisy labels during training. Current FET systems usually generate training data via crowdsourcing or distant supervision, both of which may introduce noisy labels. With the large-scale and fine-grained type sets, it is extremely difficult for humans to annotate samples accurately. The context-agnostic distant supervision method also introduces noise inevitably. Figure \ref{fig:noisylabels} shows some noisy labels from the training sets of Ultra-fine and Ontonotes. Previous denoising work in FET usually relies on pre-defined hierarchical type structures or manually labeled subsets to identify noisy labels. However, these auxiliary resources may be unavailable in real-world applications, which limits the application of such methods.

To automatically correct noisy labels without the help of auxiliary resources, an intuitive way is to treat the potentially noisy labels as unlabeled data and train the FET model with the rest clean labels. Then we can relabel candidate noisy labels to generate a cleaner dataset.
Intuitively, excluding the potentially noisy labels during training makes the model more robust, and learning from clean samples helps the model to make correct predictions on similar but mislabeled samples.
The main challenge is to identify the potentially noisy labels precisely. This problem also receives widespread attention in robust learning from noisy labels. Recent studies reveal that deep neural networks (DNNs) are prone to first learn simple patterns before overfitting noisy labels \cite{arpit2017closer}. Based on this memorization effect, there are two main approaches to filter noisy labels. Methods like Co-teaching \cite{han2018co} treat the labels with the top k\% loss as possible noisy labels (k\% is the estimated noise rate). Some other methods like SELF \cite{nguyen2020self} record the model output on different training epochs and filter noisy labels progressively.
Despite their success in other fields, those methods are difficult to directly apply to FET tasks for two reasons: i) Firstly, the noise rate can differ across FET types. Manually estimating the noise rate for each type is time-consuming. ii) Secondly, existing FET datasets usually provide a massive training set annotated by distant supervision. Current FET models can even converge before iterating through all data. Thus the multi-round methods are infeasible in practice.

To this end, we propose a novel approach to select the potentially noisy labels for FET. Figure \ref{fig:logit} briefly demonstrates the selection process.
We note the model does not strictly separate positive and negative labels before overfitting, which means the model is ``ambiguous'' on some labels in the early stage of training.
We argue this phenomenon is partially caused by noisy labels. Due to the existence of noisy labels, samples belonging to the same pattern may have different annotations. This inconsistency between annotation and underlying patterns brings difficult for the model to separate samples according to their annotations. Therefore, we propose to select potentially noisy labels by identifying the ``ambiguous labels''.
Specifically, we first fit the logits output by the model on positive and negative labels using two Gaussian distributions, respectively. Then, we calculate the posterior probability of a label being positive or negative. Finally, we select the potential noisy labels according to the agreement between original annotations and the calculated posterior probabilities.

In summary, our main contributions are three-fold:
\begin{itemize}
    \item We propose a novel method to automatically correct noisy labels in the FET training corpus without the help of auxiliary resources.
    \item We propose an effective method to automatically identify the potentially noisy labels by discriminating the ``ambiguous labels'' from the normal ones.
    \item Experiments on two widely-used benchmarks prove the effectiveness of the proposed method.
\end{itemize} 

\section{Related Work}
\subsection{Fine-grained Entity Typing}
The noisy labeling problem in FET domain has been studied for years. \cite{gillick2014context} filter noisy labels by a set of heuristics rules, \cite{ren2016afet,ren2016label} propose the partial label loss (PLL) and the partial-label embedding method to alleviate the negative influence caused by noisy labels,
\cite{onoe2019learning} utilize a small set of manually labeled training data to train a filter module and a relabel module then denoise the distantly-labeled data. \cite{chen2019improving} treat the noisy labels as unlabeled data and performs Compact Latent Space clustering (CLSC) to regularize their representation. \cite{zhang2021learning} assume each sample only has one most appropriate label and treat the underling ground truth labels as trainable parameters for fine-tuning during training. \cite{DBLP:conf/ijcai/WuZMSH21} estimate the noise confusion matrix through hierarchical label structure. Other works on the FET task dedicate to mining label dependencies \cite{xiong2019imposing,liu2021fine} designing more expressive embedding method \cite{onoe2021modeling} and exploring better automated data annotation method \cite{MLMET} to improve performance.

Most existing denoising methods in FET rely on prior auxiliary resources (pre-defined hierarchical type structures/manually labeled subsets) or assume each sample only has one most appropriate label. As compared, in this paper we consider the more restrictive setting where no auxiliary resources are available and one sample can have multiple labels simultaneously.

\subsection{Robust Learning from Noisy Labels}
Previous studies have pointed out that DNNs are susceptible to noisy labels \cite{zhang2017understanding}. Sample selection is a popular method to counteract noise, which only uses relatively clean labels to train the model. Here we briefly note several previous works in this field.

Co-teaching \cite{han2018co} simultaneously train two collaborating networks to filter out potentially noisy labels according to their loss. This method needs to control select how many small-loss instances as clean ones, which can be tricky in practice. To automatically select clean labels, DivideMix \cite{DBLP:conf/iclr/LiSH20} leveraged a one-dimensional and two-component Gaussian Mixture Model (GMM) to model the loss distribution of clean and noisy labels. SELF \cite{nguyen2020self} observed that the network's prediction is consistent on clean samples while oscillating strongly on mislabeled samples. Therefore, they select the clean labels according to the agreement between annotated labels and the running average of the network's prediction.

In practice, we found that the loss does not form a bimodal distribution in FET tasks
and fitting the GMM can not distinguish clean and noisy labels precisely. The multi-round methods are also infeasible. Training on the large-scale distantly annotated training set, current FET models can converge or even overfit noisy labels in the first epoch. To this end, we propose a novel method to select potentially noisy labels by estimating the posterior probability of a label being positive or negative according to the logits output by the model. 

\section{Methodology}
Indeed, our method corrects the noisy labels with two major steps, namely: (1) Selecting potentially noisy labels by identifying ``ambiguous labels''; (2) Re-labeling training sets by learning a robust entity typing model over the clean labels.
Next, we will first introduce the problem definition and the entity typing model used in this paper, and then present our noisy label correction method in detail.

\subsection{Problem Definition}
Entity typing datasets consist of a collection of (sentence, mention) tuples: $\mathcal{D}=\{(x_1, m_1), (x_2, m_2),..,(x_n, m_n)\}$. Given a sentence $x$ and a marked entity mention $m$, the fine-grained entity typing task aims to predict a set of types $y \subset \mathcal{T}$ of $m$, where $\mathcal{T}$ is a set of pre-defined fine-grained semantic types (\eg \emph{actor, company, building}).

\subsection{Entity Typing Model} \label{sec:model}
In FET tasks, a widely-used paradigm is to fine-tune the pre-trianed language models (PLMs) for entity typing. In this paper we adopt the prompt learning model \cite{ding2021prompt} to tune PLMs for entity typing, as it is proved to be effective for fine-tuning PLMs to specific tasks. Specifically, we reformulate the original input (sentence, mention) into the prompt input via a crafted template as follows:
\begin{equation}
    T(x, m) = x \mathrm{[P_1]} m \mathrm{[P_2] [P_3] [MASK]},
\end{equation}
where $\rm{[P_1], [P_2], [P_3]}$ are additional trainable special tokens. Their embeddings are randomly initialized. Conventionally, prompt-based tuning models need to define a set of label words $V_y$ for each $y \in \mathcal{T}$, and then the model can predict entity types by filling the $\rm{[MASK]}$:
\begin{equation}
    p(y\in \mathcal{T}|x, m)=p(\mathrm{[MASK]}=w\in V_y|T(x, m)).
\end{equation}
In this way, FET is converted to mask prediction, one of the pre-training tasks in Masked Language Models.
However, it's difficult to choose a proper label word set for each type in FET. In practice, we directly modify the decoder in the Masked Language Modeling Head (MLM Head) and let it output the probability of samples belonging to each candidate type:
\begin{equation}
    p(y\in \mathcal{T}|x, m)=\sigma(\vec{w}_y\mathrm{F}(\vec{h}_\mathrm{[MASK]}) + \vec{b}_y),
\end{equation}
where $\vec{h}_\mathrm{[MASK]}$ is $\mathrm{[MASK]}$'s contextualized representation produced by PLMs; $\mathrm{F}$ represent the Masked Language Modeling Head without decoder; $\vec{w}_y$ and $\vec{b}_y$ are trainable parameters in modified decoder. If type $y$ is in PLMs' vocabulary, we initialize $\vec{w}_y$ and $\vec{b}_y$ according to $y$'s weight and bias in MLM Head's decoder, respectively. If $y$ consists of multi tokens, we take the average of their weight and bias in MLM Head's decoder to initialize $\vec{w}_y$ and $\vec{b}_y$.

\begin{figure}[tbp]
    \centering
    \includegraphics[width=\linewidth]{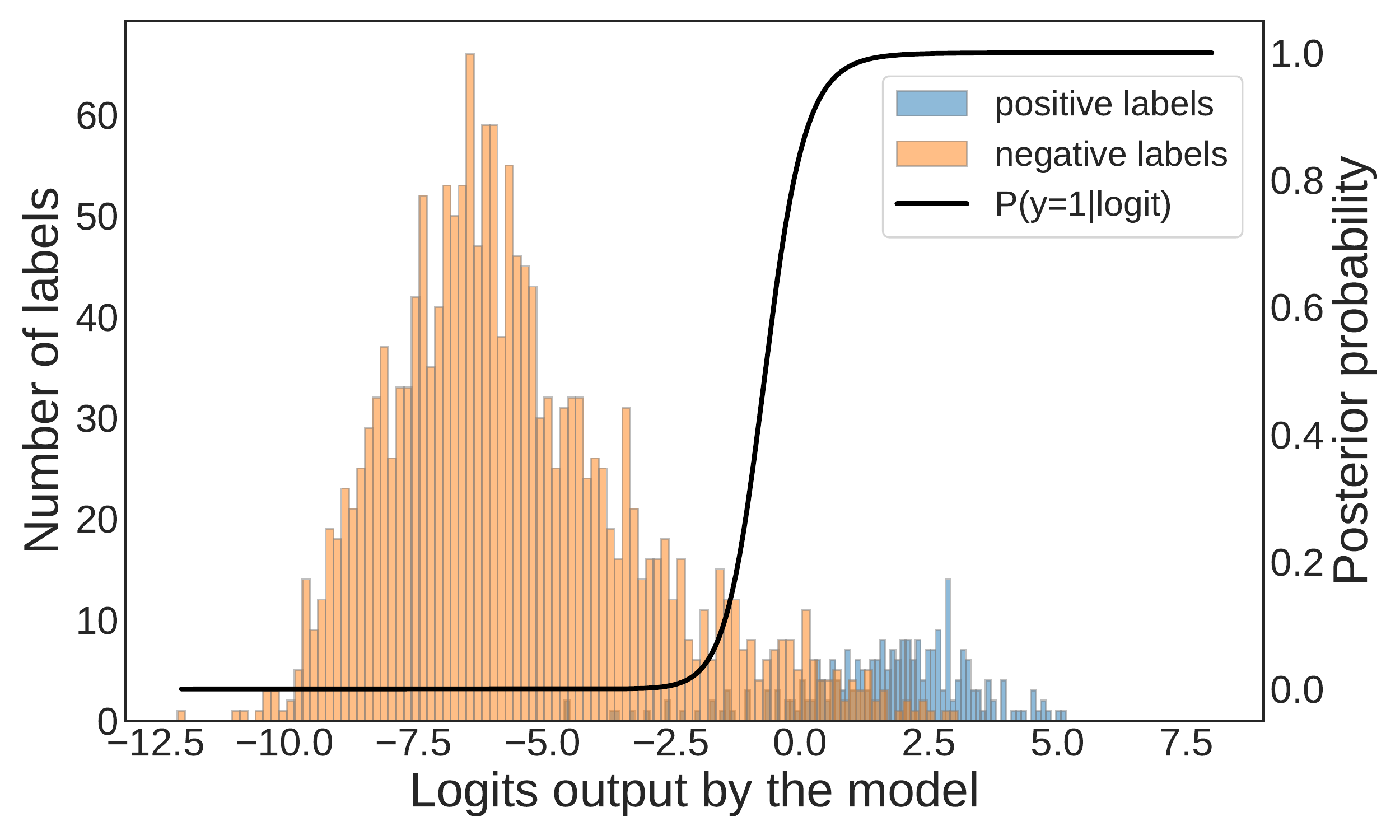}
    \caption{The logits output by the model on type ``organization'' when achieving early stopping on the Ultra-Fine dataset. Positive and negative labels are colored for illustrative purposes. This figure also shows the estimated posterior probability of a label being positive.}
    \label{fig:logit}
\end{figure}

\subsection{Candidate Noisy Labels Selection} \label{sec:noisyselection}
The basic idea of our method is to exclude potentially noisy labels from the supervision signals for robust training, then generate a cleaner dataset by relabeling the training set. In this section, we detail how to select potentially noisy labels.

According to the memorizatioin effect \cite{arpit2017closer}, deep neural networks usually memorize simple patterns before overfitting noisy labels, which is also observed in our experiments.
Figure \ref{fig:logit} shows our model's output logits on type ``organization'' when achieving early stopping on the Ultra-Fine dataset. Some samples with different annotations have similar logits, and this phenomenon is universal across types.
According to our observation, those \emph{``ambiguous''} samples mainly consists of mislabeled samples. This is because the mislabeled samples belong to the same pattern as some correctly labeled samples in datasets. Due to the memorization effect, the model tends to output similar logits on samples with the same pattern before overfitting. As a result, the predictions of noisy labels may be ``ambiguous'' or even opposite to the annotation during the early training phase.

Therefore, we propose to pre-train the model with the original annotation until early stopping, then select potentially noisy labels by independently identifying those ``ambiguous labels'' for each type. Specifically, we first estimate the posterior probability of a label being positive or negative according to the logits: $p(y=1|l)$. Then we can identify ``ambiguous labels'' through the agreement between $p(y=1|l)$ and original annotations.
During pre-training, we use the following standard binary cross-entropy (BCE) loss:
\begin{equation}
    \mathcal{L}_1=\frac{1}{|B|}\sum_{y_{s, t}\in \hat{Y}} \left[ y_{s, t}\mathrm{log}p_{s, t}+(1-y_{s, t})\mathrm{log}(1-p_{s, t}) \right],
\end{equation}
where $y_{s, t}\in \{0, 1\}$ is the original annotation in datasets; $p_{s, t}$ is the model prediction, denotes the probability of sample $s$ has type $t$; $|B|$ is the batch size. After achieving early stopping, we gather the logit output by the model and estimate the posterior probability $p(y=1|l)$:
\begin{equation}\label{eq:posterior}
    p(y=1|l) = \frac{p(l|y=1)p(y=1)}{p(l|y=1)p(y=1) +p(l|y=0)p(y=0)},
\end{equation}
where $y$ is the ground truth label and 0 means negative label, 1 means positive label; $l$ is the logit output by the model. The main challenge is estimating the prior probability: $p(y=0), p(y=1)$ and the likelihood function: $p(l|y=1), p(l|y=0)$.
We note the logits output by the model on positive (negative) labels form unimodal distribution (as shown in Figure \ref{fig:logit}). It's a natural idea to fit it using a one-dimensional Gaussian distribution:
\begin{equation}
    G(l|\mu,\delta)=\frac{1}{\sqrt{2\pi}\delta}\exp^{-\frac{(l-\mu)^2}{2\delta^2}}.
\end{equation}
However, original datasets only provide the noisy annotation $\hat{y}$ instead of the ground truth label $y$. Therefore, directly fit Gaussian distribution according to the original annotation only give $p(l|\hat{y}=0)$ and $p(l|\hat{y}=1)$. We recommend to remove those obvious noisy labels before estimating the likelihood function. Algorithm \ref{alg:filteroutlier} provides a heuristic method to filter obvious noisy labels iteratively. This method first fits a Gaussian distribution on the model output logits then removes the outliers. This process will be repeated until no outliers can be found. After getting the relative clean positive (negative) label set $Y_1$($Y_0$), the prior probability and likelihood function can be estimated in the following way:
\begin{equation}
    p(y=k)=\frac{|Y_k|}{|Y_0| + |Y_1|},
\end{equation}
\begin{equation}
    p(l|y=k)=G(l|\mu_k, \delta_k),
\end{equation}
\begin{equation}
    \mu_k=\frac{1}{|Y_k|}\sum_{y\in Y_k}l_y; \quad
    \delta_k^2=\frac{1}{|Y_k|}\sum_{y\in Y_k}(l_y-\mu_k)^2,
\end{equation}
where $k=1(0)$ denotes the positive (negative) labels. Finally, a label is selected as a potentially noisy label if it satisfies one of the following rules:
\begin{equation} \label{eq:nlselect}
    \left\{
              \begin{array}{ll}
                   \hat{y}=1 \quad &and \quad p(y=1|l_{\hat{y}})<(0.5+\epsilon)\\
                   \hat{y}=0 \quad &and \quad p(y=1|l_{\hat{y}})>(0.5-\epsilon)
              \end{array}
    \right.,
\end{equation}
where $\epsilon > 0$ is a hyperparameter, and the larger value means more labels are selected as potentially noisy labels.


\begin{algorithm}[tb]
\caption{Filter obvious noisy labels}
\label{alg:filteroutlier}
\textbf{Input}: The logits output by the model: $l$; original noisy annotation in the dataset: $\hat{Y}$. \\
\textbf{Parameter}: Outlier threshold $\alpha$. \\
\textbf{Output}: Relative clean positive (negative) label set $Y_1$ ($Y_0$)
\begin{algorithmic}[1] 
\STATE Let $Y_1=\hat{Y}_1=\{y=1|y\in \hat{Y}\}$
\STATE Let $Y_0=\hat{Y}_0=\{y=0|y\in \hat{Y}\}$
\REPEAT
\STATE Let $\hat{Y}_1 = Y_1$ and $\hat{Y}_0 = Y_0$
\STATE Calculate the average $\hat{\mu}_1$ ($\hat{\mu}_0$) and standard deviation $\hat{\delta}_1$ ($\hat{\delta}_0$) of the logits output by the model corresponding to the labels in $\hat{Y}_1$ ($\hat{Y}_0$)
\STATE Let $Y_1=\{l_y>\hat{\mu}_1-\alpha \hat{\delta}_1|y \in \hat{Y}_1\}$
\STATE Let $Y_0=\{l_y<\hat{\mu}_0+\alpha \hat{\delta}_0|y \in \hat{Y}_0\}$
\UNTIL $|Y_1| == |\hat{Y}_1|$ and $|Y_0| == |\hat{Y}_0|$
\STATE \textbf{return} $Y_1, Y_0$
\end{algorithmic}
\end{algorithm}

\subsection{Noisy Label Correction}
To correct the noisy labels in the dataset as many as possible, we propose to fine-tune the model using clean labels. Intuitively, training on clean samples helps the model to make correct predictions on those similar but mislabeled samples. For the selected potentially noisy labels, we treat them as unlabeled data and perform entropy regularization \cite{grandvalet2005semi} to reduce class overlap. Formally, the loss on a training batch during noisy label correction is as follows:
\begin{align} \label{eq:loss2}
    \mathcal{L}_2 &= \frac{1}{|B|}\sum_{y_{s, t}\in Y_c} \left[ y_{s, t}\mathrm{log}p_{s, t}+(1-y_{s, t})\mathrm{log}(1-p_{s, t}) \right] \notag \\
    & - \frac{\beta}{|B|}\sum_{y_{s, t}\in Y_n}\left[p_{s, t}\mathrm{log}p_{s, t} + (1-p_{s, t})\mathrm{log}(1-p_{s, t}) \right],
\end{align}
where $Y_c$ and $Y_n$ stand for clean and potentially noisy label set respectively; $\beta$ is a hyperparameter that balances the BCE loss and the entropy regularization. Finally, we relabel potentially noisy labels to generate cleaner datasets:
\begin{equation} \label{eq:relabel}
    \left\{
        \begin{array}{ll}
             y_{s, t}^{'} = 1,& \  if \  p_{s, t} > 0.5 \\
             y_{s, t}^{'} = 0,& \  if \  p_{s, t} < 0.5
        \end{array}
    \right. \quad for \  y_{s, t} \in Y_n,
\end{equation}
where $y_{s, t}^{'}$ is the new label after noisy label correction. Algorithm \ref{alg:correction} summarizes the whole process of our method.


\begin{algorithm}[tb]
\caption{Noisy labels correction for FET}
\label{alg:correction}
\textbf{Input}: Noisy labeled training set $\mathcal{\hat{D}}_{train}$; validation set $\mathcal{D}_{val}$.\\
\textbf{Parameter}: Threshold $\epsilon$; weight of entropy regularization $\beta$; fine-tuning step $k$ on clean samples.\\
\textbf{Output}: Cleaner training set $\mathcal{D}_{train}$.

\begin{algorithmic}[1] 
\STATE Fine-tune model $\mathcal{M}$ on $\mathcal{\hat{D}}_{train}$ until the performance drops on $\mathcal{D}_{val}$
\STATE Filter obvious noisy labels by Algorithm \ref{alg:filteroutlier}
\STATE Select potentially noisy labels by Equation (\ref{eq:nlselect})
\STATE Conduct $k$ step fine-tuning on model $\mathcal{M}$ using loss function in Equation (\ref{eq:loss2})
\STATE Relabel $\mathcal{\hat{D}}_{train}$ according to Equation (\ref{eq:relabel}) to generate $\mathcal{D}_{train}$
\STATE \textbf{return} $\mathcal{D}_{train}$
\end{algorithmic}
\end{algorithm} 

\section{Experiments}
\subsection{Datasets and Experiment Setup}
\paragraph{Datasets.} We conduct experiments on two standard fine-grained entity typing datasets: Ultra-fine and OntoNotes. Followed \cite{liu2021fine} we use the 2k/2k/2k train/dev/test splits for Ultra-fine which contains 6K manually annotated samples, 2519 categories. Original OntoNotes dataset contains 25k/2k/9k train/dev/test data, 89 categories. \cite{choi2018ultra} offers an augmented training set containing 0.8M samples. We conduct experiments on both versions. In terms of evaluation metrics, we follow the widely used setting proposed in \cite{ling2012fine}. On Ultra-Fine, we report macro-averaged precision, recall, and F1-score. On OntoNotes, we report accuracy, macro-averaged F1-score, and micro-averaged F1-score.

\paragraph{Baselines.} For ultra-fine entity typing task, we compare with the following approaches: 
\begin{itemize}
    \item \textbf{UFET} \cite{choi2018ultra} performs multi-label classification on features encode by GloVe+LSTM and a character level CNN. 
    \item \textbf{LabelGCN} \cite{xiong2019imposing} models the label dependency using a graph propagation layer.
    \item \textbf{LDET} \cite{onoe2019learning} performs relabeling and sample filtering to denoise the datasets before training FET models.
    \item \textbf{MLMET} \cite{MLMET} automatically generates ultra-fine entity typing labels to train the entity typing model by combining hypernym extraction patterns with a masked language model.
    \item \textbf{Box} \cite{onoe2021modeling} proposes to use box embeddings instead of vector embeddings to capture latent type hierarchies.
    \item \textbf{LRN} \cite{liu2021fine} learns and reasons label dependencies in both sequence-to-set and ene-to-end manner.
\end{itemize}
For the OntoNotes dataset, in addition to the baselines mentioned above, we also compare with several approaches focusing on noisy labeling problems introduced in related work \cite{chen2019improving,zhang2021learning,DBLP:conf/ijcai/WuZMSH21}.

\paragraph{Implementation.}
We use BERT-base-cased as backbone structures, AdamW optimizer with the learning rate of 2e-5 on Ultra-fine and 2e-6 on OntoNotes. The training batch size is 16 for all datasets. Other hyper-parameters are tuned by grid search and the optima configurations are $\{\epsilon=0.1, \alpha=2.0, \beta=0.5, k=2000\}$ on Ultra-Fine; $\{\epsilon=0.4, \alpha=3.0, \beta=2.0, k=5000\}$ on augmented OntoNotes; $\{\epsilon=0.3, \alpha=2.0, \beta=1.0, k=5000\}$ on original OntoNotes.

\subsection{Overall Results}
\begin{table}[t]
  \centering
    \begin{tabular}{lccc}
    \toprule
    \multicolumn{1}{c}{\textbf{Method}} & \textbf{P} & \textbf{R} & \textbf{F1} \\
    \midrule
    UFET  & 47.1  & 24.2  & 32.0  \\
    LabelGCN & 50.3  & 29.2  & 36.9  \\
    LDET  & 51.5  & 33.0  & 40.2  \\
    Box   & 52.8  & 38.8  & 44.8  \\
    LRN  & 54.5  & 38.9  & 45.4  \\
    MLMET & 53.6  & 45.3 & 49.1  \\
    \midrule
    Ours (original) & 58.2$\pm$0.8  & 40.8$\pm$0.1 & 48.0$\pm$0.2  \\
    Ours (denoised) & 55.6$\pm$0.4 & 44.7$\pm$0.3 & \textbf{49.5$\pm$0.1} \\
    \quad w/o filter & \textbf{59.6$\pm$0.3} & 40.8$\pm$0.3 & 48.5$\pm$0.3 \\
    \quad w/o ER & 53.0$\pm$0.2 & \textbf{46.2$\pm$0.3} & 49.3$\pm$0.1 \\
    \bottomrule
    \end{tabular}%
  \caption{Macro-averaged Precision, Recall, and F1 of different approaches on Ultra-Fine test set.}
  \label{tab:UltraFine}%
\end{table}%

\begin{table}[t]
  \centering
  \resizebox{\linewidth}{!}{
    \begin{tabular}{lccc}
    \toprule
    \multicolumn{1}{c}{\textbf{Method}} & \textbf{Acc} & \textbf{Macro F1} & \textbf{Micro F1} \\
    \midrule
    \multicolumn{4}{c}{\textbf{with augmentation}} \\
    \midrule
    UFET  & 59.5  & 76.8  & 71.8  \\
    LabelGCN & 59.6  & 77.8  & 72.2  \\
    LDET  & 64.9  & 84.5  & 79.2  \\
    LRN   & 64.5  & 84.5  & 79.3  \\
    MLMET & 67.4  & 85.4  & 80.4  \\
    \midrule
    Ours (original) &  63.7$\pm$0.3  &  84.8$\pm$0.2  & 79.6$\pm$0.4 \\
    Ours (denoised) &  \textbf{67.8$\pm$0.1}  &  \textbf{87.1$\pm$0.2}  & \textbf{81.5$\pm$0.1} \\
    \quad w/o filter &  67.3$\pm$0.4 & 87.0$\pm$0.1  & 81.3$\pm$0.1 \\
    \quad w/o ER &  65.3$\pm$0.3  & 86.2$\pm$0.3 & 81.0$\pm$0.2 \\
    \midrule
    \multicolumn{4}{c}{\textbf{without augmentation}} \\
    \midrule
    LRN & 56.6 & 77.6 & 71.8 \\
    NFETC-CLSC$_{hier}$ & 62.8$\pm$0.3 & 77.8$\pm$0.4 & 72.0$\pm$0.4 \\
    NFETC-AR$_{hier}$ & \textbf{64.0$\pm$0.3} & 78.8$\pm$0.3 & 73.0$\pm$0.3 \\
    FBTree & \textbf{64.0$\pm$0.1} & 78.4$\pm$0.2 & 72.5$\pm$0.2 \\
    \midrule
    Ours (original) & 52.6$\pm$2.2 & 77.1$\pm$1.6 & 71.1$\pm$1.2 \\
    Ours (denoised) & 59.2$\pm$0.2 & \textbf{81.3$\pm$0.3} & \textbf{75.3$\pm$0.4} \\
    \quad w/o filter & 59.6$\pm$0.6 & 79.8$\pm$0.3 & 74.6$\pm$0.3 \\
    \quad w/o ER & 56.9$\pm$0.3  &  80.3$\pm$0.3  &  73.8$\pm$0.2 \\
    \bottomrule
    \end{tabular}%
  }
  \caption{Strict accuracy, macro-averaged F1, and micro-averaged F1 of different approaches on OntoNotes test set.}
  \label{tab:OntoNotes}%
\end{table}%

Table \ref{tab:UltraFine} and Table \ref{tab:OntoNotes} show the overall results on Ultra-Fine and OntoNotes test sets. We report the performance of our model (described in Section \ref{sec:model}) training on the original training set (original) and the denoised training set (denoised) respectively. We can see that: i) Our noisy label correction method is effective across datasets. After using the denoised training set generated by our method, the model performance significantly improved and outperform other advanced baselines on almost all metrics; ii) Generate cleaner training data is critical to FET. Both our method and MLMET generate cleaner data to train the FET model. We can see they significantly outperform other baselines. iii) Prompt-learning is an effective approach to leverage PLMs in FET. Compared with methods using vanilla fine-tuning (\eg Box, LRN), prompt-leaning brings significant performance gains. Training on the original noisy training data, our prompt-learning model still achieves competitive performance against other baselines. 

Compared with previous works focusing on noisy labeling problems in FET, our method is more universal. NFETC-CLSC$_{hier}$, NFETC-AR$_{hier}$ and FBTree require pre-defined hierarchical type structures during training. They are difficult to apply to datasets like Ultra-Fine whose type set is not being explicitly hierarchical. LDET requires a small set of manually labeled training data, which is not provided by datasets like OntoNotes. In contrast, our method corrects noisy labels without auxiliary resources, which can be applied to all FET datasets. It's also worth noting that our approach is orthogonal to most existing FET models. In practice, our method can combine with other advanced FET models to yield better performance.

\subsection{Ablation Study}
Filter obvious noisy labels (filter) can estimate the posterior probability $p(y=1|l)$ more accurately, which is critical to potentially noisy label selection. Entropy regularization (ER) encourages decision boundaries to fall in low-density regions, which is helpful for noisy label correction. To better understand their effect on the model, we conduct the ablation study on Ultra-Fine and OntoNotes datasets. The results are reported in Table \ref{tab:UltraFine} and Table \ref{tab:OntoNotes}.

\paragraph{Effect of Filter Obvious Noisy Labels.} This strategy brings improvements to all datasets, especially in the Ultra-Fine dataset. Without filtering obvious noisy labels, our method only brings 0.5\% improvement on Macro-averaged F1 compared to training on the original training set. This is mainly due to the long tail problem in Ultra-Fine: many long-tail types only have several positive samples. So the outliers (obvious noisy labels) can seriously affect the estimation of likelihood function $p(l|y=1)$, lead to a poor estimation of $p(y=1|l)$. In practice, we recommend enabling this strategy in datasets with long-tail problems.

\paragraph{Effect of Entropy Regularization.} Entropy regularization brings improvements to all datasets. This indicates that treating potentially noisy labels as unlabeled data and training the model in a semi-supervised manner helps with noisy label correction.

\subsection{Case Study}
\begin{figure}[t]
    \centering
    \includegraphics[width=\linewidth]{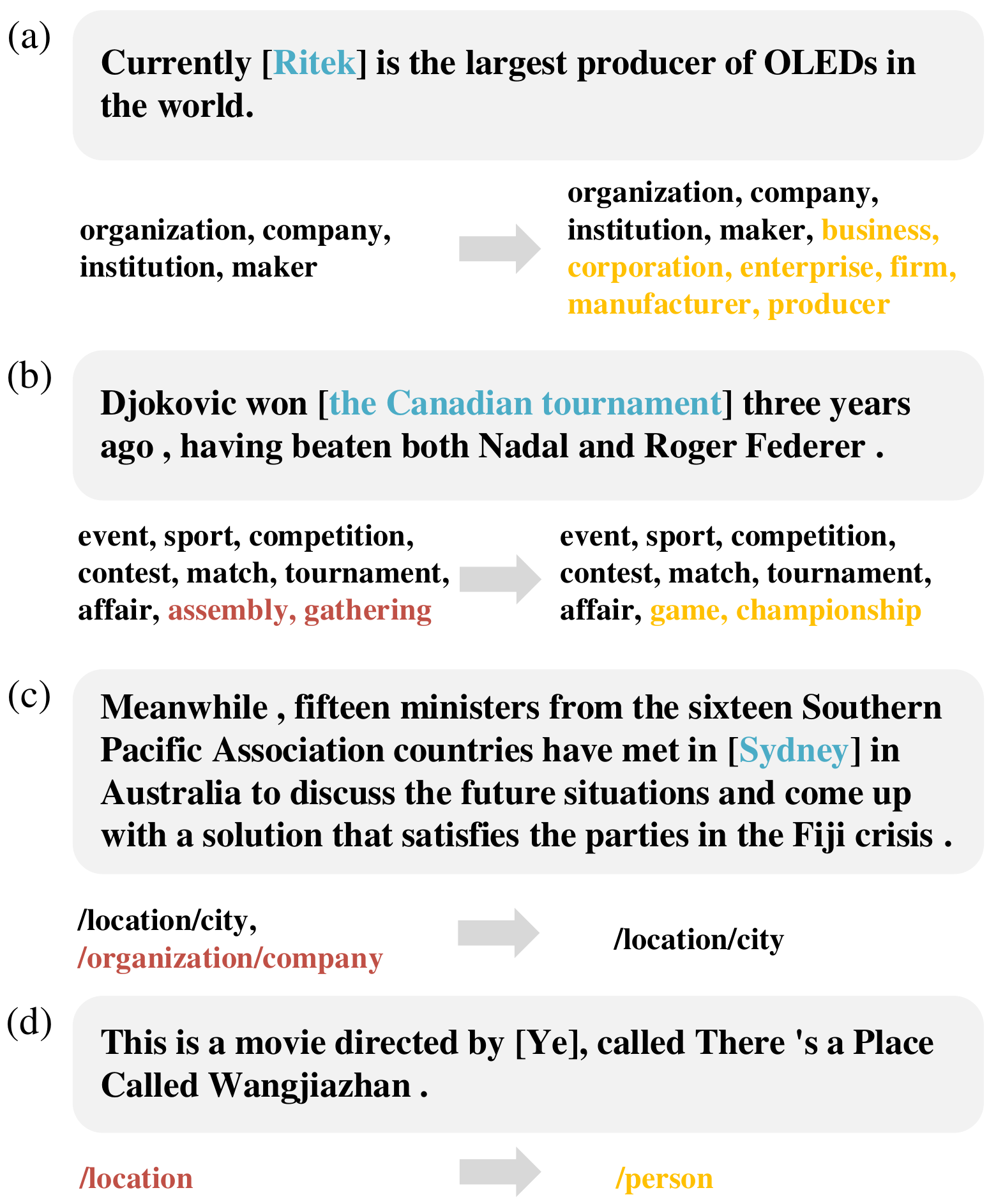}
    \caption{Example of noisy labels (left) and the denoised labels (right). Example (a), (b) are manually-annotated samples taken from Ultra-Fine; (c), (d) are distantly-labeled samples taken from OntoNotes. False positive labels are colored in red, false negative labels are colored in yellow.}
    \label{fig:case}
\end{figure}

Figure \ref{fig:case} shows examples of the original noisy labels and the denoised labels produced by our method. In example (a), the original human-annotated labels fail to cover all proper types. Our method generates a more comprehensive type set. In example (b), our method removes labels that are not highly correlated with this sample (assembly, gathering) and supplements two fine-grained labels (games, championship). In example (c), our method retains the correct label ``/location/city'' and removes the false positive label ``/organization/company''. In example (d), our method reassigns the correct label for this sample.

\subsection{Visualization}
\begin{figure}[t]
    \centering
    \subfigure[Original]{\label{fig:tsne_1}
        \begin{minipage}{0.47\linewidth}
            \centering
            \includegraphics[width=\linewidth]{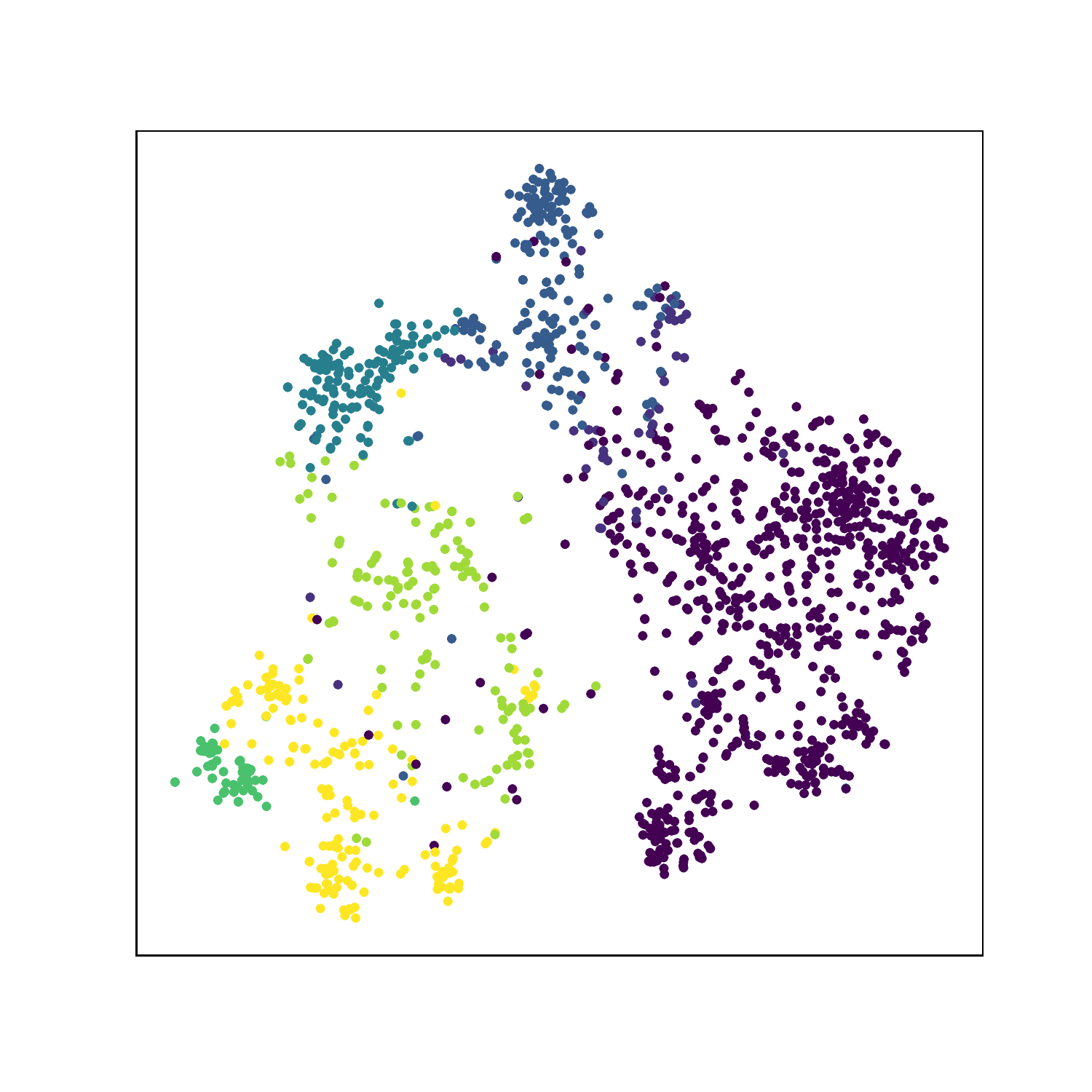}
        \end{minipage}
    }
    \subfigure[Denoised]{\label{fig:tsne_2}
        \begin{minipage}{0.47\linewidth}
            \centering
            \includegraphics[width=\linewidth]{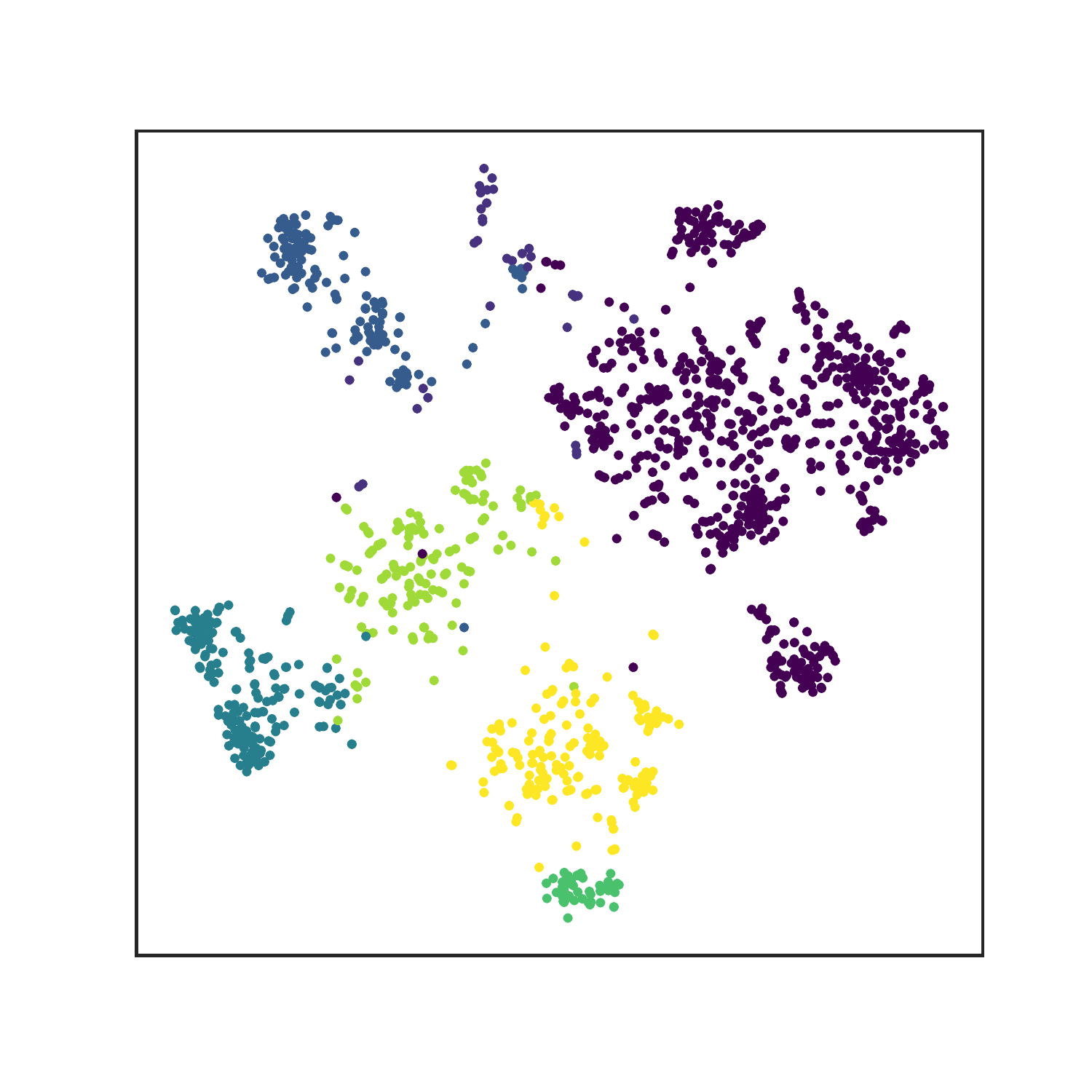}
        \end{minipage}
    }
    \caption{T-SNE virtualization of mention embeddings on the original (left) and denoised (right) training set of Ultra-Fine when achieving early stopping. Samples are shown in different colors according to their coarse-grained types.}
    \label{fig:tsne}
\end{figure}

To better illustrate how noisy labels affect model performance in FET, we train the model using the original and denoised training data separately and visualize mention embeddings in Figure \ref{fig:tsne}\footnote{Some mentions may have multiple coarse-grained types, we only visualize mentions with one coarse-grained type.}. 

In Figure \ref{fig:tsne_1}, there is overlap between different type clusters. Intuitively, noisy labels in training sets prevent the model from learning a clear decision boundary for each type. 
\cite{AUM} offers an explanation for this. Take a mislabeled sample for example. The gradient updates from those similar but correctly labeled samples encourage the model to predict the underlying ground truth through generalization. However, the gradient update from this mislabeled sample itself forces the model to memorize the incorrectly assigned label. These two opposing gradient updates result in a poor margin between different types. Consequently, the model is easy to make incorrect predictions for samples near decision boundaries.
From Figure \ref{fig:tsne_2} we can see the margin between different types increases and the decision boundaries are clear. This phenomenon shows correcting the noisy labels brings more discriminative sample representation in FET, and our method indeed generates cleaner training data.

\section{Conclusion}
In this paper, we focus on the noisy labeling problem in the fine-grained entity typing task and propose a novel approach to automatically correct the noisy labels. Different from previous denoise approaches, our method does not require auxiliary resources, which is more universal. Experiments on two benchmark datasets prove the effectiveness of our method.

\section*{Acknowledgments}
This work was supported in part by the National Natural Science Foundation of China under Grant No.61602197, Grant No.L1924068, Grant No.61772076, in part by CCF-AFSG Research Fund under Grant No.RF20210005, in part by the fund of Joint Laboratory of HUST and Pingan Property \& Casualty Research (HPL), and in part by the National Research Foundation (NRF) of Singapore under its AI Singapore Programme (AISG Award No: AISG-GC-2019-003). Any opinions, findings and conclusions or recommendations expressed in this material are those of the authors and do not reflect the views of National Research Foundation, Singapore. The authors would also like to thank the anonymous reviewers for their comments on improving the quality of this paper.

\bibliographystyle{named}
\bibliography{ijcai22}

\end{document}